\documentclass[10pt,twocolumn,letterpaper]{article}

\usepackage[pagenumbers]{wacv} % To force page numbers, e.g. for an arXiv version

\usepackage{graphicx}
\usepackage{amsmath}
\usepackage{amssymb}
\usepackage{booktabs}

\usepackage[pagebackref,breaklinks,colorlinks]{hyperref}

\usepackage[capitalize]{cleveref}
\usepackage[accsupp]{axessibility}
\crefname{section}{Sec.}{Secs.}
\Crefname{section}{Section}{Sections}
\Crefname{table}{Table}{Tables}
\crefname{table}{Tab.}{Tabs.}

\usepackage{tikz}
\usepackage{comment}
\usepackage{color}

\usepackage{booktabs}
\usepackage{algorithm}
\usepackage{algorithmic}
\usepackage{multirow}
\usepackage{tabularx}
\usepackage{verbatim}
\usepackage{subcaption}
\usepackage{pifont}
\newcommand{\cmark}{\ding{51}}
\newcommand{\xmark}{\ding{55}}

\newcommand{\revise}[1]{{#1}}

\usepackage[capitalize]{cleveref}
\crefname{section}{Sec.}{Secs.}
\Crefname{section}{Section}{Sections}
\Crefname{table}{Table}{Tables}
\crefname{table}{Tab.}{Tabs.}

\def\wacvPaperID{808} % *** Enter the WACV Paper ID here
\def\confName{WACV}
\def\confYear{2024}

\begin{document}

\title{Video Instance Matting}

\author{ Jiachen Li\textsuperscript{1}, Roberto Henschel\textsuperscript{2}, Vidit Goel\textsuperscript{2}, Marianna Ohanyan\textsuperscript{2}, Shant Navasardyan\textsuperscript{2}, Humphrey Shi\textsuperscript{1,2} \\
{\small \textsuperscript{1}SHI Labs $@$ Georgia Tech \& Oregon \& UIUC,   \textsuperscript{2}Picsart AI Research (PAIR)}\\
}

\maketitle

%%%%%%%%% ABSTRACT
\begin{abstract}
Conventional video matting outputs one alpha matte for all instances appearing in a video frame so that individual instances are not distinguished. 
While video instance segmentation provides time-consistent instance masks, results are unsatisfactory for matting applications, especially due to applied binarization.
To remedy this deficiency, we propose \textbf{Video Instance Matting~(VIM)}, that is, estimating alpha mattes of each instance at each frame of a video sequence. To tackle this challenging problem, we present \textbf{MSG-VIM}, a Mask Sequence Guided Video Instance Matting neural network, as a novel baseline model for VIM. MSG-VIM leverages a mixture of mask augmentations to make predictions robust to inaccurate and inconsistent mask guidance. It incorporates temporal mask and temporal feature guidance to improve the temporal consistency of alpha matte predictions. 
Furthermore, we build a new benchmark for VIM, called \textbf{VIM50}, which comprises 50 video clips with multiple human instances as foreground objects. To evaluate performances on the VIM task, we introduce a suitable metric called Video Instance-aware Matting Quality~(VIMQ).
Our proposed model MSG-VIM sets a strong baseline on the VIM50 benchmark and outperforms existing methods by a large margin.
The project is open-sourced at \href{https://github.com/SHI-Labs/VIM}{https://github.com/SHI-Labs/VIM}. 

\end{abstract}

\vspace{-2mm}
%%%%%%%%% BODY TEXT
\section{Introduction}
%%%%
% P1: motivation: previous solutions are not enough 
% P2: introduce video instance matting
% P3: benchmark -> metric -> MGVIM
% Summary
% Fig1 VIS vs VM vs VIM
%%%%
\vspace{-1mm}
\begin{figure}[tb]
\centering
\includegraphics[width=0.5\textwidth]{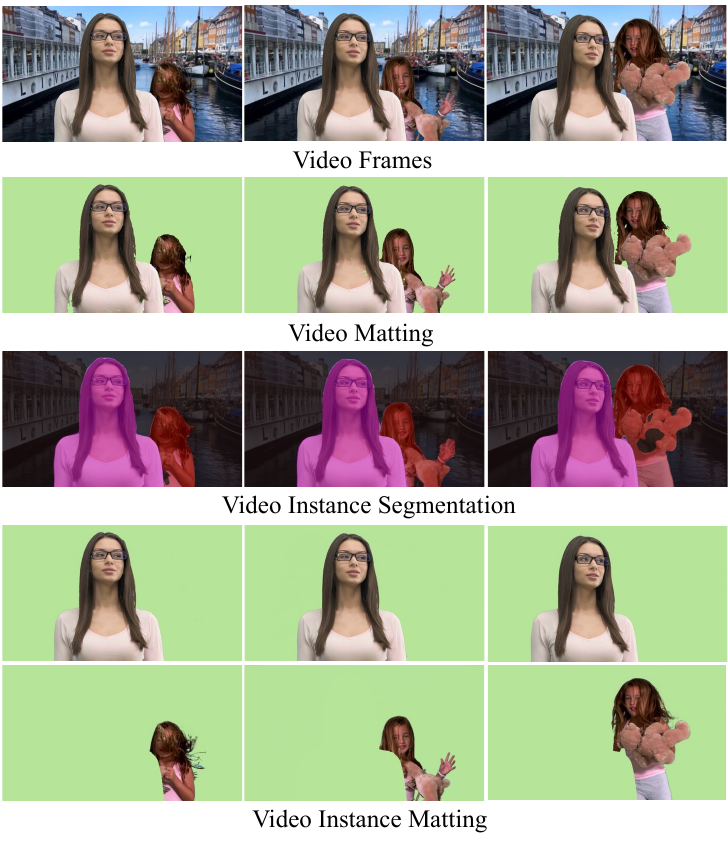}
\vspace{-5mm}
\caption{\textbf{\revise{Video Instance Matting compared with related tasks.}}
Depicted are three frames of the VIM50 benchmark. The results of Video Matting, Video Instance Segmentation, and Video Instance Matting are obtained from RVM~\cite{lin2021robust}, SeqFormer~\cite{wu2021seqformer}, and MSG-VIM, respectively. Video Instance Matting is the task of recognizing and tracking each foreground instance, then estimating the alpha matte of the corresponding instance at each frame of the video sequence.
}
\vspace{-6mm}
\label{fig:teaser}
\end{figure}

Recently, video matting has drawn much attention from industry and academia as it is widely used in video editing and video conferencing~\cite{lin2021robust}. 
Current deep learning based video matting methods~\cite{ke2020green,lin2021robust,sengupta2020background,lin2021real,li2022vmformer} output a single alpha matte for all instances appearing in the foreground for each frame. However, different applications require sequences of alpha mattes \textit{and} the separation into the instances. 
%Consider the scenario of an online meeting with multiple people in front of a camera. If we want to blend over everything except for the speaker with a virtual background, we need for the entire recording the alpha mattes of the speaker. Especially in cases of partial occlusion, this is not solved by current video matting methods.
One approach towards instance-aware video matting is to employ video instance segmentation~\cite{voigtlaender2019mots, kim2020video, athar2020stem, bertasius2020classifying, cao2020sipmask}, which segments and tracks each object instance appearing in a video sequence. 
Unfortunately, the generated masks are binary and coarse at the outline of an instance, making them inadequate for high-quality instance-aware video editing. 
Several image matting works~\cite{yu2021mask, sun2022human} have thus focused on converting segmentation masks into alpha mattes by adopting mask guidance from instance segmentation. However, such approaches do not track instances and are thus not applicable to video editing.
In summary, there is no off-the-shelf solution for high-quality instance-aware video matting.
%\JC{As a result, there is no off-the-shelf solutions that meet the needs of high-quality instance-aware video matting.}
%

Motivated by these observations, we extend video matting to a  multi-instance scenario in Section \ref{VIM_DEF}, called  \textbf{Video Instance Matting~(VIM)}, a new task aiming at estimating alpha mattes of each instance at each frame given a video sequence, as shown in Figure~\ref{fig:teaser}.
To tackle this task, we propose in Section \ref{sec:method} a new baseline method, called \textit{Mask Sequence Guided Video Instance Matting} (\textbf{MSG-VIM}), which takes mask sequences from a video instance segmentation (VIS) method as guidance and transforms them into time-consistent, high-quality alpha matte sequences.
In more detail, we employ a VIS method to obtain a sequence of coarse binary masks for each instance.  
Then, masks are concatenated with corresponding video frames and passed to an encoder-decoder-based model, which returns a sequence of high-quality alpha mattes for each instance.
This mask-guided architecture allows our method to benefit from future advances in VIS \revise{\textit{without} re-training the MSG-VIM model.}
Furthermore, we propose a mixture of mask augmentations during training, which make MSG-VIM less susceptible to error propagation from mask sequence guidance caused by the employed VIS method.
We then apply temporal mask guidance~(TMG) and temporal feature guidance~(TFG) to the model, incorporating temporal information for the alpha matte creation. Accordingly, MSG-VIM can compensate for individual incorrect input masks, leading to improved matting quality.

In order to evaluate the performance of MSG-VIM and other methods on the VIM task, we establish in Section \ref{VIM_DEF} \textbf{VIM50}, a benchmark for VIM, which comprises 50 video clips with multiple human instances as foregrounds for evaluation. 
We further propose the \textit{Video Instance-aware Matting Quality}~(VIMQ) metric to evaluate video instance matting performance. It simultaneously considers recognition, tracking, and matting quality. 
We compare MSG-VIM to video matting, video instance segmentation, and image instance matting methods on the proposed VIM50 benchmark. The experiments show that the VIM task is not sufficiently well-handled by existing approaches, justifying the focus on this challenging task. Moreover, the experiments demonstrate that carefully incorporating mask sequence guidance and temporal modeling is crucial to obtain accurate results. 
As shown in Figure~\ref{fig:teaser}, MSG-VIM shows not only better instance-level matting quality, but also conventional video matting quality if we merge all instance mattes.

To summarize, our contributions are as follows:
\begin{itemize}
    \item We propose Video Instance Matting, a new task aiming at predicting alpha mattes of each foreground instance at each frame given a video sequence as input.
    
    \item We establish a benchmark for the proposed VIM, called VIM50, which comprises 50 videos with multiple human instances as foregrounds. Furthermore, we propose VIMQ as an evaluation metric for VIM.
    
    \item We propose MSG-VIM, a Mask Sequence Guided Video Instance Matting network, as a simple and strong baseline model for the VIM50 benchmark. %, \JC{which can be further applied to video matting task.}
\end{itemize}

\section{Related Works}
\vspace{-1mm}
\subsection{Image Matting}
\vspace{-1mm}
%\RHT{Add references to image matting. It is actually a very old technique, see \url{https://en.wikipedia.org/wiki/Matte_(filmmaking)}. So there are actually way more approaches than what you listed as "previous solutions".}

Image matting~\cite{wang2008image}, \ie, the overlay of a foreground image with a background image, is an important technique in photography and filmmaking and a classical problem in computer vision.  
To tackle the task, several methods have focused on detecting the transition areas of the alpha mattes using low-level features~\cite{aksoy2017designing,chuang2001bayesian,bai2007geodesic,chen2013knn,grady2005random}. Recently, deep learning based methods have been proposed that tackle image matting end-to-end using guidance from a manually-created  trimap~\cite{xu2017deep,wang2018deep,zhu2017fast,li2020natural,yu2020high,qiao2020attention,yu2021cascade,forte2020f}. 
Applicability has been simplified by exploring trimap-free approaches~\cite{chen2018semantic}, \eg, using coarse segmentation masks~\cite{liu2020boosting, yu2021mask, yaman2021alpha, chen2022pp, li2023matting} as guidance.
To eliminate the drawback of image matting methods, which output exactly one alpha matte for an image, HIM~\cite{sun2022human} proposes multi-instance matting. To this end, instance-level masks are generated from MaskRCNN~\cite{he2017mask} and further refined by incorporating corresponding image data, resulting in instance-level alpha mattes. In contrast to VIM, HIM works on a frame-level so that the sequence of alpha mattes per instance is not provided. We show in our experiments that simply connecting the frame-wise results of HIM is not sufficient to obtain time-consistent high-quality results.

\subsection{Video Matting}
\vspace{-1mm}
Compared to image matting, the task of video matting is to estimate sequences of alpha mattes in a video sequence. By leveraging temporal context the quality of predictions improves. Trimap-based methods add spatial-temporal feature aggregation~\cite{sun2021deep,zhang2021attention} to improve the accuracy and consistency of alpha matte predictions at each frame. Most trimap-free solutions use a trimap~\cite{seong2022one} only for the first frame, or background images~\cite{sengupta2020background,lin2021real}. MODNet~\cite{ke2020green}, RVM~\cite{lin2021robust} and VideoMatt~\cite{li2023videomatt} directly predict mattes from a video. VMFormer~\cite{li2022vmformer} adopts the transformer to solve the video matting task. Yet, these methods output only one alpha matte per frame, which covers all foreground instances. When used in a multi-instance scenario, these video matting methods are incapable of distinguishing alpha mattes for different instances. In contrast, VIM methods output instance-aware mattes, which is crucial for various video editing applications such as instance-selective human removal in videos.

%\RHT{explain: Why this is bad!! Mention something like this is exactly where we "fill the gap"}

%Compared to image matting, the task of video matting is to estimate alpha mattes from the input of a video clip. Previous trimap-based methods add spatial-temporal feature aggregation~\cite{sun2021deep,zhang2021attention} to improve the temporal consistency of predictions. Trimap-free solutions use background images~\cite{sengupta2020background,lin2021real} or trimap~\cite{seong2022one} as auxiliary inputs only for the first frame. MODNet~\cite{ke2020green} and RVM~\cite{lin2021robust} take video clips as inputs without any auxiliary parts. However, these video matting methods cannot distinguish alpha matte of each instance, making them infeasible to the task of video instance matting.

\subsection{Video Instance Segmentation}
\vspace{-1mm}
The goal of video instance segmentation~(VIS) is to simultaneously perform detection, tracking and segmentation of all instances appearing in a video sequence. The baseline approach MaskTrackRCNN~\cite{yang2019video} adds a tracking head to Mask RCNN~\cite{he2017mask} to achieve tracking ability. 
Subsequent works~\cite{voigtlaender2019mots, kim2020video, athar2020stem, bertasius2020classifying, cao2020sipmask, goel2021msn} have progressively improved the performance with better representation learning and unified architectures such as transformers~\cite{wang2021end,cheng2021mask2former,wu2021seqformer, jain2021semask, jain2023oneformer}. 
Yet, resulting masks are not suitable for matting tasks as (i) they are too coarse at the outlines of instances and (ii) they are binary.

%\RHT{Again what is meant with "recognize"?}
%
%Video instance segmentation~(VIS) is proposed to track, recognize and segment each instance in a video clip. The baseline approach MaskTrackRCNN~\cite{yang2019video} adds a tracking head upon Mask RCNN~\cite{he2017mask} to improve the tracking ability on the detected instance. Subsequent works~\cite{voigtlaender2019mots, kim2020video, athar2020stem, bertasius2020classifying, cao2020sipmask} have progressively improved the performance with better representation learning and unified architecture. Recently, transformer-based solutions~\cite{wang2021end,cheng2021mask2former,wu2021seqformer} are introduced that take inputs of the whole video clips and output a series of mask predictions for each instance. Nevertheless, the binary mask predictions are too coarse at the border of instances, which are infeasible for high-quality video editing on instances.

\section{Video Instance Matting}
\vspace{-1mm}
\label{VIM_DEF}
%%% -Problem definition 
%%% -VIM50 benchmark & fig benchmark 
%%% -Evaluation Metric
%%%%
%%%%
\begin{table}[tb]
%\resizebox{0.5\textwidth}{!}{%
\centering
\begin{tabular}{c|ccc}
Task  &Alpha Matte &Instance \\ \hline
Video Instance Segmentation &\xmark &\cmark \\ \hline
Video Matting &\cmark &\xmark \\  \hline  
Video Instance Matting  &\cmark  &\cmark  \\ \hline
\end{tabular}
\vspace{-1mm}
\caption{High-level comparison between video instance matting and related tasks.} 
\vspace{-5mm}
\label{vim50}
\end{table}

\subsection{Problem Definition}
\vspace{-1mm}
We consider a video sequence $\mathbf{I} \in \mathbb{R}^{T \times H \times W \times 3}$ with $T$ frames, where each frame $I_{t}, t \in \{1,\ldots,T\}$ has spatial dimension $H \times W$. The conventional alpha matting~\cite{wang2008image} is defined for a single image $I_{t}$. The task is to find a composition into a foreground image $F_{t}$ and background image $B_{t}$ together with an alpha matte $\alpha_{t} \in [0,1]^{H \times W \times 1}$, \ie , 
\begin{equation}
\vspace{-1mm}
\label{eqn:image_matte}
 I_{t} = \alpha_t \circ F_t + (\mathbf{1} - \alpha_t) \circ B_t.
 \vspace{-1mm}
\end{equation}
Here, $\circ$ denotes the Hadamard product and $\mathbf{1}$ is the all-one matrix of appropriate dimension. Our proposed video instance matting task extends image-based matting to sequences of mattings for multiple instances.  
Hence, assuming that $N$ instances appear in $\mathbf{I}$, the task is to find alpha mattes $\alpha_{t}^{i} \in [0,1]^{H \times W \times 1}$ for all $i \in \{1,\ldots,N\}$, for all $t \in \{1,\ldots,T\}$ such that
\begin{equation}
\vspace{-1mm}
 I_t = \sum_{i=1}^N  \alpha_t^i \circ F_t^i + (\mathbf{1} - \sum_{i=1}^N\alpha_t^i) \circ B_t,
 \vspace{-1mm}
\end{equation}
which is a natural extension of \eqref{eqn:image_matte} by assuming that 
\begin{equation}
\vspace{-1mm}
 F_t = \sum_{i=1}^N F^i_t,  \; \alpha_t = \sum_{i=1}^N \alpha_t^i, \; \alpha_t^i \circ F_t^j = 0 \, \forall i \neq j.
\vspace{-1mm}
\end{equation}
Thus, the alpha mattes and foreground images of an image $I_{t}$ are dissected, according to the appearing instances.
A video instance matting method thus has to recognize and track each instance, and to estimate the alpha matte of the corresponding instance at each frame. Compared to (i) video matting and (ii) video instance segmentation, it requires (i) instance-level alpha mattes and (ii) accurate mask predictions without binarization. We show a high-level comparisons between video instance matting and other tasks in Table~\ref{vim50}. 

\subsection{VIM50 Benchmark}
\vspace{-1mm}
\label{ssec:vim50}
There is no benchmark that provides real-world video clips with instance-level alpha matte annotations for each frame. To be able to validate VIM approaches, we thus create a new benchmark for evaluation. To this end, we leverage instance-level foregrounds and backgrounds contained in existing single-instance matting datasets to form our multi-instance video matting benchmark VIM50. Specifically, we use VideoMatte240k~\cite{lin2021real}, which provides high-resolution foreground human instances with alpha matte annotations, and the background video dataset DVM~\cite{sun2021deep} to composite the VIM50 benchmark.
We randomly select 50 consecutive background frames and select $N$ times a clip of 50 consecutive frames from the foreground sequences, each showing a single human. We save corresponding backgrounds, foregrounds, and alpha mattes.
%Eventually, 
We composite 50 testing clips of resolution $1920 \times 1080$, each consisting of $50$ frames. VIM50 comprises 35,10 and 5 video clips with two, three and four human instances, respectively, to cover different levels of crowdedness. 35 video clips depict partially occluded persons, thus posing challenges inputs regarding tracking and matting quality. Exemplary frames of VIM50 are presented in Figure~\ref{fig:teaser} and in the Appendix~\ref{appendix:benchmark}.
%\RH{TODO: Add reference to figure or section in appendix.}
%%%% add a comparison table
%%%% to-do-list: evaluation of the VIM50
%\RH{TODO: Detailed dataset evaluation!}
%\RHT{We should definitely show and reference samples of VIM in the main paper. Also show in the appendix which data we provide (all mattes and forground masks)}
%\JC{What kind of samples we want to show here, just video frames or video frames + instance-level GT (like the one in the Appendix?)}
%\RH{Actually, Figure 1 is enough for the main paper. So lets reference Fig1 here, and just present more in the appendix. }

\subsection{Evaluation Metric}
\vspace{-1mm}
We propose a new metric, which we term \textit{Video Instance-aware Matting Quality}~(VIMQ), to evaluate the proposed video instance matting task. It combines the instance-level matting quality (MQ)~\cite{sun2022human}, tracking quality~\cite{voigtlaender2019mots} (TQ), and recognition quality~\cite{kirillov2019panoptic} (RQ) via
\begin{equation}
\vspace{-1mm}
    \mathrm{VIMQ = RQ \cdot TQ \cdot MQ}.
\vspace{-1mm}
\end{equation}
The metrics require a minimum-cost maximal matching between predicted and ground truth instance-sequences of alpha mattes. To this end, we apply the Hungarian algorithm~\cite{cheng2021mask2former} on sequence-averaged L1 distances. The true positive set $TP$ comprises the matched sequences of alpha mattes $(\alpha_{1}^{i},\ldots ,\alpha_{T}^{i})$, whose  average intersection over union between the binarized masks $[\alpha^{i}] := ([\alpha_{1}^{i}],\ldots ,[\alpha_{T}^{i}])$ and  corresponding binarized ground truth masks  $[\Bar{\alpha}^{i}] := ([\Bar{\alpha}_{1}^{i}],\ldots ,[\Bar{\alpha}_{T}^{i}])$ is above $\rho$. Binarization is done via $\alpha>0$. 
%to find an one-to-one matching between prediction $\boldsymbol\alpha = (\alpha_1, \alpha_2, ... \alpha_T )$ and ground truth $\Bar{\boldsymbol\alpha} = (\Bar{\alpha}_1, \Bar{\alpha}_2, ... \Bar{\alpha}_T )$, based on the clip-level cost matrix.  
We denote by $N_{TP}$, $N_{FP}$ and $N_{FN}$ the number of corresponding true positives, false positives and false negatives, respectively. 
Finally, we compute for each true positive instance its frame-averaged intersection over union to the ground truth masks and obtain 
$\mathrm{RQ}$, an IoU-weighted F1 score of the alpha mattes:
\begin{equation}
\vspace{-1mm}
    \mathrm{RQ} = \frac{\sum_{i \in {TP}}\mathrm{IoU([\alpha^{i}],[\Bar{\alpha}^{i}])}}{N_{TP}} \cdot \frac{ N_{TP} }{N_{TP}  + \frac{1}{2} N_{FP} + \frac{1}{2} N_{FN}}.
\vspace{-1mm}
\end{equation}
%%% instance-level to instance-aware
%We propose a new metric, which we term Video instance-level Matting Quality~(VMQ), to evaluate the proposed video instance matting task. It takes instance-level accuracy of alpha matte predictions~\cite{sun2022human}, tracking ability~\cite{voigtlaender2019mots}, and recognition quality~\cite{kirillov2019panoptic} at the same time into consideration. It is formulated as
%\begin{equation}
%    \mathrm{VMQ = RQ \times TQ \times MQ} .
%\end{equation}
%%% \alpha_1 includes multiple \alpha^1_1, ...
%%%% short explain how Hungarian works with examples
%RQ, TQ, and MQ stand for Recognition Quality, Tracking Quality, and Matting Quality, respectively. The first step is to apply clip-level one-to-one matching between alpha mattes predictions of all instances $\boldsymbol\alpha = \{\alpha_1, \alpha_2, ... \alpha_T \}$ and ground truth $\Bar{\boldsymbol\alpha} = \{\Bar{\alpha}_1, \Bar{\alpha}_2, ... \Bar{\alpha}_T \}$ based on the Hungarian algorithm~\cite{cheng2021mask2former}. \RH{What if number of predicted instance does not equal number of ground truth instances?} Then, we compute the Recognition Quality~(RQ), which is an IoU-weighted F1 score  
%\begin{equation}
%    \mathrm{RQ} = \frac{\mathrm{IoU_{TP}}}{N_{TP}} \times \frac{N_{TP}}{N_{TP} + \frac{1}{2} N_{FP} + \frac{1}{2} N_{FN}}.
%\end{equation}
%\RH{Definitions $N_{TP},N_{FP},N_{FN}$ used in the equation missing} 
To measure the Tracking Quality~(TQ), we compute a frame-wise minimum-cost maximal matching between TP and ground truth. A deviating assignment compared to the sequence-wise matching is counted as an ID switch error, since it fails to track the corresponding instance at the respective frame. 
The tracking quality is thus defined as
\begin{equation}
\vspace{-1mm}
    \mathrm{TQ} = 1 -  \frac{\sum_{i \in {TP}} \sum_{t=1}^T \mathrm{IDS}(i,t)}{N_{TP} \cdot T},
\vspace{-1mm}
\end{equation}
where $\mathrm{IDS}(i,t) = 1$ in the case of an ID switch error for $i$ at frame $t$, and $0$ otherwise.   
In order to evaluate the matting quality, we compare the ground truth alpha matte $\Bar{\alpha}^i_t$ with prediction $\alpha^i_t$ using a similarity metric $S$ defined by
\begin{equation}
\vspace{-1mm}
    S(\Bar{\alpha}^i_t, \alpha^i_t)) =  1 - \min(1, \omega \xi(\Bar{\alpha}^i_t, \alpha^i_t)) \in [0,1].
\vspace{-1mm}
\end{equation}
$\omega$ is a manually set weight and $\xi$ computes the mean distance between true positive predicted and ground-truth mattes, inspired by IMQ~\cite{sun2022human}. We utilize Mean Absolute Difference~(MAD), Mean Squared Error~(MSE), and direct temporal gradients on Sum of Squared Differences~(dtSSD)~\cite{erofeev2015perceptually} as metrics. 
MSE and MAD are used to evaluate the frame-level accuracy, while dtSSD shows temporal consistency of the estimations. Finally, we introduce the Matting Quality (MQ) as 
\begin{equation}
\vspace{-1mm}
    \mathrm{MQ} =  \frac{\sum_{i \in {TP}}\sum_{t=1}^T S(\Bar{\alpha}^i_t, \alpha^i_t)}{ N_{TP}  \cdot T}.
\vspace{-1mm}
\end{equation}
We chose $\rho = 0.5$, $\omega = 50$ for VIMQ$_{\mathrm{mse}}$ and VIMQ$_{\mathrm{mad}}$, and $\omega = 10$ for VIMQ$_{\mathrm{dtssd}}$ in all experiments.

\section{Method}
\label{sec:method}
\vspace{-1mm}
%%% -MSG-VIM & Fig2
%%% -MMA
%%% -TG
%%% -Others
\begin{figure*}[tb]
\centering
\includegraphics[width=1.0\textwidth]{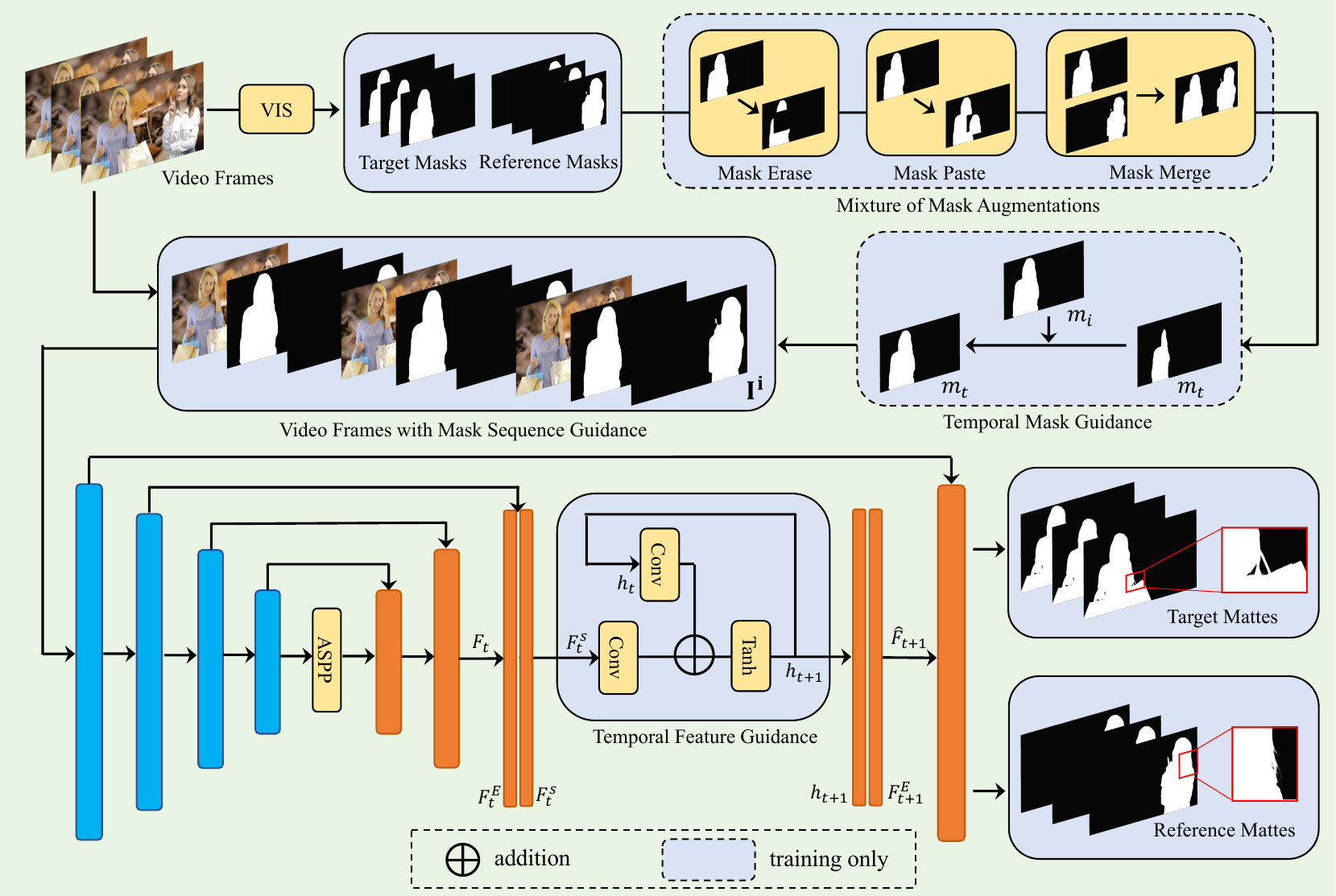}
\caption{\textbf{The Architecture of MSG-VIM.} A VIS method is used to obtain mask sequences of target and reference instances. The sequences of mask guidance are concatenated with video frames as inputs to the matting network, which outputs the refined alpha matte predictions of both target and reference instances. Details of the MSG-VIM are described in Section~\ref{sec:method} and illustrated in the \revise{Appendix~\ref{appendix:model_architecture}.}}
\vspace{-4mm}
\label{fig:architecture}
\end{figure*}

%This section introduces our approach \textit{Mask Sequence Guided Video Instance Matting }~(MSG-VIM) for video instance matting. %To improve the accuracy of predictions, we present a mixture of mask augmentations, which are employed during training to increase robustness to erroneous mask guidance. %Furthermore, 
%We introduce temporal guidance to masks and feature maps to improve temporal consistency of the predictions. Finally, we introduce the loss function used during training and explain the inference step.
\subsection{MSG-VIM}
\label{ssec:msg-vim}
To demonstrate feasibility of the VIM task, we propose a baseline model. To this end, we take advantage of the progress made in video instance segmentation and use their mask sequence predictions as auxiliary input. The subsequent network converts them into alpha mattes, thus focusing on improving matting quality.
Specifically, given a video clip $\mathbf{I} \in \mathbb{R}^{T \times H \times W \times 3}$ showing  $N$ human instances, we employ a VIS method to obtain mask predictions $\mathbf{m}^i = (m^i_1, \ldots , m^i_T)$ for instance $i \in N $, where $m^{i}_{t}$ is the prediction for instance $i$ at frame $t$.  To enable the matting model to focus on improving the matting result, we split the whole mask sequences into two groups: (i) $\mathbf{m}^i_{\mathbf{tar}} := \mathbf{m}^i$ and (ii) $\mathbf{m}^i_{\mathbf{ref}} := \sum_{j \ne i}^N \mathbf{m}^j$. Reference masks have been proven to reduce false positive predictions of non-selected  instances~\cite{sun2022human}, which help the subsequent matting model to focus on refining the target mask. Then, we apply a mixture of mask-oriented augmentation in Section~\ref{ssec:mix_aug} to make the model robust to inaccurate and misleading mask guidance. We introduce temporal guidance in Section~\ref{ssec:temporal_guidance}, which enables the model to exploit matte signals across frames, leading to improved results.   Finally, target and reference masks are concatenated with $\mathbf{I}$, resulting in 
%After the split of mask sequence guidance, we concatenate them to the channels of video frames $\mathbf{I}$ to form for instance $i \in N$ the input
\begin{equation}
\label{eqn:input_concat}
\vspace{-1mm}
    \mathbf{I}^i = \mathrm{Concat}( \mathbf{I} ,   \mathbf{m_{tar}}^{i}, \mathbf{m_{ref}}^{i}  ) \in \mathbb{R}^{T \times H \times W \times 5}.
\vspace{-1mm}
\end{equation}
We utilize a neural network $U$, which takes  $\mathbf{I}^{i}$ as input. It uses an encoder-decoder design to estimate alpha mattes of the target and reference instances as shown in Figure~\ref{fig:architecture}. 
The encoder is based on ResNet34~\cite{li2020natural}, which extracts sequences of feature maps at multiple resolutions. Then, the feature maps with the lowest resolution are sent to an Atrous Spatial Pyramid Pooling~(ASPP) ~\cite{chen2017deeplab} module and upsampled to higher resolutions in the decoder. We also build skip connections between the feature maps of the same resolution at the encoder and the decoder. 
For the prediction of the alpha mattes of the target and reference instances, we adopt Progressive Refinement Module~\cite{yu2021mask}, and add a light convolution layer upon the  feature maps at the decoder to obtain the target and reference mattes. To exploit temporal context and improve time-consistency of the predictions, we add temporal feature guidance between consecutive frames with a lightweight recurrent neural network, see Section~\ref{ssec:temporal_guidance}. Finally, the alpha matte sequences $\mathbf{\boldsymbol\alpha_{tar}}^{i}$ and $\mathbf{\boldsymbol\alpha_{ref}}^{i}$ of instance $i$ are predicted jointly from the upsampled feature maps at the decoder of $U$:
\begin{equation}
\vspace{-1mm}
 (\mathbf{\boldsymbol\alpha_{tar}}^{i}, \mathbf{\boldsymbol\alpha_{ref}}^{i}) := U (\mathbf{I}^{i} ).
 \vspace{-1mm}
\end{equation}

\subsection{Mixture of Mask Augmentations}
\vspace{-1mm}
\label{ssec:mix_aug}
%\RH{This part is not clear to me. Why didnt you save for each sequence the instance segmentation results from a VIS method and used them as input? You are writing you could not do that. 
%}
%\RH{Did you also train on real instance segmentation masks?} No
%\JC{Use Mixture of ...}
%\RHT{I find the naming strange. You want to express that you use several different augmentation methods right? "Mixture of Mask Augmentations"?}

During training, we randomly choose foregrounds and backgrounds and composite training clips on-the-fly. To keep training efficient, we do not use VIS inference to obtain mask guidance.  Instead, we convert a labeled alpha matte $\alpha$ into a mask $m_{\alpha}$ via binarization. To mimic mask sequence guidance from a VIS model, we randomly apply erosion and dilation to the binarized mattes before creating the inputs for  \eqref{eqn:input_concat}.
% \RHT{Mask Erase: Do you apply this to all frames, or also randomly picking frames?} 
% \RHT{Mask Swap: I dont understand. You are swapping within mask regions? So you will have the same mask but different foreground images? If you have two instances do you swap between different masks? I understood from your last explanation you swap with background.}
% \RH{We swap within one mask image $[\alpha]$ two regions,}
%\RHT{I dont understand, where is randomness here? You merge the entire target mask with reference mask, and you do this for all frames. So it would be just one mask for all frames. Also what is target mask after merging? What is reference after merging?} 
Still, resulting masks are often more accurate than typical results from a  VIS model. Considering that mask sequence guidance can be inaccurate during inference, we apply a mixture of mask-oriented clip-level augmentations during training to make the MSG-VIM model robust to such guidance, as shown in Figure~\ref{fig:architecture}. The first strategy is to use Mask Erase, \ie  randomly erasing parts of the mask guidance at each frame during training. Then, we adopt Mask Paste that randomly selects two mask regions, and paste one region to the other one at each frame to add perturbations to the input. We further apply Mask Merge, which randomly picks frames and merges the whole target and reference mask at the selected frame to make the model robust to joint predictions of different instances. With the proposed mask augmentation strategy, the model becomes robust to errors induced by inaccurate mask guidance. We observe significant performance improvements especially on RQ and MQ, as shown in Table~\ref{ablation_mma}.

\subsection{Temporal Guidance}
\vspace{-1mm}
\label{ssec:temporal_guidance}
Using temporal information during matte creation is expected to improve quality. To this end, MSG-VIM utilizes temporal guidance \wrt mask sequences and feature maps.

\noindent \textbf{Temporal Mask Guidance}
We introduce a temporal mask guidance to make the model robust to individual localisation errors caused by the video instance segmentation method. To this end, during training, we consider a mask sequence $\mathbf{m} = (m_1, \ldots , m_T)$. For each  $m_t$ at frame $t$, we randomly choose $ i \in  \{1,\ldots, T\}$ and merge the corresponding masks, \ie we set $m_t := m_t + m_{i}$. 
%We introduce a temporal mask guidance to make the model robust to spurious localisation errors caused by the video instance segmentation method. To this end, during training, we consider a mask sequence $\mathbf{m} = (m_1, \ldots , m_T)$. For each  $m_t$ at frame $t$, we randomly choose $ i \in  \{1,\ldots, T-t+1\}$ \RHT{a small $i \in T$?} \JC{change $m_{t+i}$ to $m_i $} and merge the corresponding masks, \ie we set $m_t := m_t + m_{t+i} + m_{t-i}$. 
%
%and select the mask $m_{t \pm i}$ from frame $t \pm i$. 
%Then, we merge it with $m_t := m_t + m_{t \pm i}$ ,
Our experiments show that MQ, TQ and RQ are improved with temporal mask guidance, as shown in Table~\ref{ablation_msgvim}.

\noindent \textbf{Temporal Feature Guidance}
We further add a lightweight convolutional recurrent network at the second highest feature map resolution at the decoder %of the matting network 
to exploit temporal information, see Figure~\ref{fig:architecture}. For timestep t, the feature map, which we denote by $F_{t}$, is split into $F_{t}^{S}$ and  $F_{t}^{E}$ by the first and second half of the channels. We compute  the recurrent state $h_{t+1}$, which is used in the next iteration to add temporal context to $F_{t+1}$,
The hidden state is given via $h_{t+1} = \mathrm{tanh} ( \mathrm{Conv}(h_t) + \mathrm{Conv}(F_t^{S}) )$.
%
%\begin{equation}
%    h_{t+1} = \mathrm{Tanh} ( \mathrm{Conv}(h_t) + \mathrm{Conv}(F_t^{S}) ).
%\end{equation}
%}
The update is performed via $\hat{F}_{t+1} := \mathrm{Concat}(h_{t+1},F_{t+1}^{E})$.
%Now $h_{t+1}$ is concatenated with the feature map of first encoder layer for timestep $t+1$ to form the final feature map for the next frame $t+1$. 
With the temporal feature guidance, the model can exploit temporal context so that the predictions can benefit from it. As shown in Table~\ref{ablation_tg},  it further improves the temporal consistency of the predictions across frames.
%\RHT{Please check if description of Temporal feature guidance is correct}
\subsection{Loss Function}
%\noindent \textbf{Loss Function} 
We apply a loss function to the predictions of target alpha matte $\mathbf{\boldsymbol\alpha_{tar}}$ and reference alpha mattes $\mathbf{\boldsymbol\alpha_{ref}}$ simultaneously. For frame $t$, we use L1 loss $L_\alpha^{t}$, pyramid Laplacian loss $L_{lap}^{t}$, and composition loss  $L_{com}^{t}$~\cite{lin2021robust,hou2019context,sun2021deep,sun2022human}. The composition loss at frame $t$ is defined as 
\begin{equation}
\vspace{-1mm}
    L_{com}^{t} = \| \sum_{i=1}^N \alpha^i_t \circ F^i_t + (\mathbf{1} - \sum_{i=1}^N  \alpha^i_t) \circ B_t - I_t \|_1.
\vspace{-1mm}
\end{equation}
The total loss averages the losses over all frames:
\begin{equation}
\vspace{-1mm}
    L = T^{-1} \sum_{t=1}^T  L_\alpha^{t} + L_{lap}^{t} + L_{com}^{t}.
\vspace{-1mm}
\end{equation}
%%% add \sum_t for loss 
%\RHT{Loss divided by T?}

\subsection{Inference}
For inference, we apply a VIS method to the video sequence and obtain mask sequences of each instance. We then run inference iteratively for each instance $i \in N$, forming $\mathbf{m_{tar}}^{i}$ and $\mathbf{m_{ref}}^{i}$, sending them to the MSG-VIM network, which output the alpha mattes $\mathbf{\boldsymbol\alpha_{tar}}^{i}$ and $\mathbf{\boldsymbol\alpha_{ref}}^{i}$ for instance $i$. Finally, all $\mathbf{\boldsymbol\alpha_{tar}}^{i}$ are merged as final outputs.

%Inference starts by applying a VIS method to the input sequence. 

%We then form the mask sequence guidance $\mathbf{m_{tar}}^{i}$ and $\mathbf{m_{ref}}^{i}$ for instance $i$.  We then run inference iteratively for each instance $i \in N$ with the corresponding target masks $\mathbf{m_{tar}}^{i}$ and reference masks $\mathbf{m_{ref}}^{i}$ as auxiliary inputs. Then, the matting networks output the refined alpha mattes predictions $\mathbf{\boldsymbol\alpha^i_{tar}}$ and $\mathbf{\boldsymbol\alpha^i_{ref}}$. We add the alpha matte predictions of all target instances into the final set $\mathbf{\boldsymbol\alpha_{tar}}$ for performance evaluation.

\section{Experiments}
\vspace{-1mm}
\label{sec:experiments}
%%% Dataset & Training & Testing 
%%% Ablation on modules
%%% SoTA comparisons
%%% Other benchmarks
\begin{table}[tb]
\centering
\resizebox{0.5\textwidth}{!}{
\begin{tabular}{>{\centering}m{6mm}>{\centering}m{6mm}>{\centering}m{6mm}|llll}
MMA & TMG & TFG &  RQ$\uparrow$ & TQ$\uparrow$ & MQ$_{\mathrm{mse}}$$\uparrow$ & VIMQ$_{\mathrm{mse}}$$\uparrow$ \\ \hline
- &- &- &63.28 &91.32 &42.86 &24.77 \\
\checkmark  &- &-  &70.92 &92.16 &51.91 &33.92 (\textcolor{cyan}{+9.15})  \\
\checkmark  &\checkmark &- &71.67 &92.55 &55.81 &37.02 (\textcolor{cyan}{+12.25})  \\
\checkmark  &\checkmark  &\checkmark  &\textbf{72.72} &\textbf{93.17} &\textbf{56.52} &\textbf{38.29} (\textcolor{cyan}{+13.52})  \\ \hline
\end{tabular}}
\vspace{-2mm}
\caption{Ablation on the MSG-VIM Setup. MMA: Mixture of Mask Augmentations, TMG: Temporal Mask Guidance, TFG: Temporal Feature Guidance.} %The mask sequence guidance is from MaskTrackRCNN~\cite{yang2019video}, and the models are tested on the VIM50 benchmark.} %\textbf{Bold} numbers indicate best performance among all models.}
\vspace{-3mm}
\label{ablation_msgvim}
\end{table}
\begin{table}[tb]
\centering
\resizebox{0.5\textwidth}{!}{
\begin{tabular}{l|llll}
Method &  RQ$\uparrow$ & TQ$\uparrow$ & MQ$_{\mathrm{mse}}$ $\uparrow$ & VIMQ$_{\mathrm{mse}}$ $\uparrow$ \\ \hline
MMA  &\textbf{70.92} &\textbf{92.16} &\textbf{51.91} &\textbf{33.92} \\
$-$ Mask Erase &61.64 &91.70 &47.87 &27.06 (\textcolor{cyan}{-6.86}) \\
$-$ Mask Merge &62.15 &92.71 &48.32 &27.84 (\textcolor{cyan}{-6.08}) \\ 
$-$ Mask Paste &64.06 &92.05 &50.97 &31.62 (\textcolor{cyan}{-2.30}) \\ \hline
\end{tabular}}
\vspace{-2mm}
\caption{Ablation on the Mixture of Mask Augmentations.} %All models are tested on the VIM50 benchmark. \textbf{Bold} numbers indicate best performance among all models.} 
\vspace{-4mm}
\label{ablation_mma}
\end{table}
\begin{table}[tb]
\centering
\resizebox{0.5\textwidth}{!}{
\begin{tabular}{l|llll}
Method &  RQ$\uparrow$ & TQ$\uparrow$ & MQ$_{\mathrm{dtssd}}$ $\uparrow$ & VIMQ$_{\mathrm{dtssd}}$ $\uparrow$ \\ \hline
MMA &70.92 &92.16 &24.92 &16.28 \\
+ TMG &71.67 &92.55 &27.76 &18.41 (\textcolor{cyan}{+2.13}) \\
+ TFG  &\textbf{72.72} &\textbf{93.17} &\textbf{28.51} &\textbf{19.32} (\textcolor{cyan}{+3.04}) \\ \hline
\end{tabular}}
\vspace{-2mm}
\caption{Ablation on the Temporal Guidance. We select \textit{dtSSD} as the similarity metric to highlight the improvement in the temporal consistency of the predictions.} %We apply Temporal Mask Guidance~(TMG) and Temporal Feature Guidance~(TFG) to the baseline model.} %All models are tested on the VIM50 benchmark. \textbf{Bold} numbers indicate best performance among all models.} 
\label{ablation_tg}
\vspace{-4mm}
\end{table}

%In this section, we introduce details regarding implementation, training and testing of MSG-VIM, followed by ablation studies. We then compare MSG-VIM with methods from related tasks on the VIM50 benchmark. Finally, we show the applicability of MSG-VIM for conventional video matting and discuss qualitative results. 

\subsection{Implementation Details}
\vspace{-1mm}
\noindent \textbf{Datasets} Since there is no video instance matting dataset, we select foregrounds and background data from different benchmarks separately to train our model. 
For the foregrounds, we choose high-resolution clip-level instances from the remaining part of VideoMatte240k~\cite{lin2021real}, which excludes the human instances used in VIM50. For the backgrounds, we choose clip-level video backgrounds without human instances from the remaining part of DVM~\cite{sun2021deep},  excluding the backgrounds used in VIM50. Also, we select 20,000 frame-level image backgrounds from BG20k~\cite{li2022bridging} to make the model robust to diverse environments. During training, we select two to four instances as foregrounds and iteratively add them to the video backgrounds~\cite{sun2022human} to composite the training data, which follows the same practice we used when compositing the VIM50 benchmark.\\ 
\noindent \textbf{Training Setting} During training, we use eight RTX A6000 GPUs with two video clips per GPU as bath size, each containing 10 consecutive frames. We empoy the Adam optimizer with $\beta_1 = 0.5$ and $\beta_2 = 0.99$. We use cosine learning rate decay with an initial learning rate of  $1e^{-3}$. Training lasts for $2 \cdot 10^{4}$ iterations, with warm-up at the first $2 \cdot 10^{3}$ iterations. For the data augmentation, we first generate mask sequence guidance for each instance from the corresponding alpha matte sequence, by binarization followed by random erosion and dilation. We separately apply RandomAffine and RandomCrop to the foregrounds, masks, alpha mattes, and backgrounds. Then, they are composited iteratively and concatenated to train the matting network, see Section \ref{ssec:mix_aug}. 

\begin{figure}[tb]
\centering
 \includegraphics[width=0.48\textwidth]{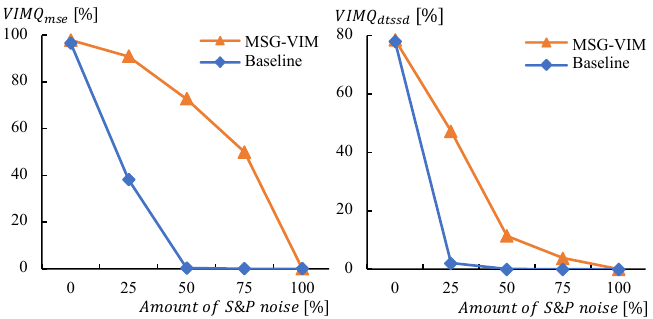}
\vspace{-3mm}
\caption{Performance of MSG-VIM and the baseline model \wrt VIMQ metrics on VIM50.} %The x-axis represents the fraction of pixels of the mask guidance to which Salt-and-Pepper noise is applied. The y-axis represents the performance \wrt the VIMQ metrics.}
%MSG-VIM shows better robustness towards the  salt-and-pepper noise over the baseline model of Section~\ref{ssec:msg-vim}.} % \RHT{x-axis label: Fraction of pixels with additional Salt-and-Pepper noise} \RHT{Remove Title}}
\vspace{-5mm}
\label{fig:vis_mask_per}
\end{figure}

\begin{table*}[tb]
%\small
\centering
\resizebox{1.0\textwidth}{!}{%
\begin{tabular}{c|cc|cc|cc|cc|cc}
 Method &Backbone &FPS &RQ $\uparrow$  &TQ $\uparrow$ &MQ$_{\mathrm{mse}}$ $\uparrow$ &VIMQ$_{\mathrm{mse}}$ $\uparrow$ &MQ$_{\mathrm{mad}} \uparrow$ &VIMQ$_{\mathrm{mad}}$ $\uparrow$ &MQ$_{\mathrm{dtssd}}$ $\uparrow$ &VIMQ$_{\mathrm{dtssd}}$ $\uparrow$\\ \hline \hline
\multicolumn{1}{l}{\revise{\textit{Video Instance Segmentation}}} \\ \hline 
MTRCNN~\cite{yang2019video}  &ResNet50 &24.5  &37.43 &78.00 &15.49 &4.52 &8.45 &2.47 &2.21 &0.65 \\
SeqFormer~\cite{wu2021seqformer} &ResNet50 &75.7  &79.83 &98.01 &36.01 &28.18 &23.69 &18.53 &4.13 &3.23 \\ \hline \hline
\multicolumn{1}{l}{\revise{\textit{Video Matting}}} \\ \hline
MODNet~\cite{ke2022modnet}  &MobileNetV3 &124.0  &32.67 &65.52 &18.72 &4.01 &13.02 &2.79 &7.85 &1.68 \\
RVM~\cite{lin2021robust}   &MobileNetV3 &131.5  &37.29 &74.44 &20.87 &5.79 &14.43 &4.00 &5.00 &1.39 \\
\hline \hline
\multicolumn{1}{l}{\revise{\textit{Mask-Guided Image Matting}}} \\ \hline
MGMatting$^*$~\cite{yu2021mask}  &\revise{ResNet34-UNet} &31.4  &56.08 &87.39 &26.88 &13.17 &17.74 &8.69 &14.16 &6.94 \\
MGMatting$^\dagger$~\cite{yu2021mask} &\revise{ResNet34-UNet} &31.4   &70.83 &96.49 &43.13 &29.48 &27.84 &19.03 &25.33 &17.31  \\
InstMatt$^*$~\cite{sun2022human}  &\revise{ResNet34-UNet} &27.2  &65.63 &92.63 &42.26 &25.69 &30.45 &18.51 &21.19 &12.88 \\ 
InstMatt$^\dagger$~\cite{sun2022human}  &\revise{ResNet34-UNet} &27.2  &82.57 &97.88 &64.39 &52.04 &47.94 &38.74 &33.32 &26.93 \\\hline \hline

\multicolumn{1}{l}{\revise{\textit{Video Instance Matting}}} \\ \hline
MSG-VIM$^*$ &\revise{ResNet34-UNet} &30.7 &72.72 &93.17 &56.52 &38.29 &40.49 &27.43 &28.51 &19.32 \\
MSG-VIM$^\dagger$ &\revise{ResNet34-UNet} &30.7 &\textbf{91.21} &\textbf{98.34} &\textbf{78.87} &\textbf{70.74} &\textbf{59.60} &\textbf{53.46} &\textbf{46.40} &\textbf{41.62} \\\hline \hline
\end{tabular}}
%}
\vspace{-2mm}
\caption{Performance of SOTA methods on the VIM50 benchmark. Models with $^*$ and $^\dagger$ use mask guidance from MTRCNN and SeqFormer, separately. 
%\JC{The table shows that (i) VIS, VM and Mask-Guided IM methods are not feasible solutions to the VIM task. (ii) MSG-VIM outperforms these SOTA methods and sets a strong baseline on the VIM50 benchmark.} 
}
\vspace{-4mm}
\label{sota_iccv}
\end{table*} 
\begin{table}[tb]
\centering
\begin{tabular}{c|ccc}
Model &MAD$\downarrow$  &  MSE$\downarrow$  &  Grad$\downarrow$   \\ \hline %\midrule[1.5pt]  %dtSSD\\ \hline %\midrule[1.5pt]
BGMv2~\cite{lin2021real} &20.35 &14.26 &22.79 \\ %&3.08 \\
MODNet~\cite{sun2021modnet} &11.13 &5.54 &15.30 \\ %&3.08 \\
RVM~\cite{lin2021robust}  &6.57 &1.93 &10.55 \\ %&1.90 \\
MSG-VIM  &\textbf{6.47} &\textbf{1.73} &\textbf{10.40} \\%&6.59 \\ %&2.41 \\
\end{tabular}
\vspace{-2mm}
\caption{Performance on the video matting benchmark of RVM~\cite{lin2021robust}. MSG-VIM is evaluated without retraining.} %$\dagger$BGMv2 was tested by us since it is not reported in RVM~\cite{lin2021robust}. MSG-VIM is evaluated without retraining the model.}
\label{rvm_hr_testet}
\vspace{-4mm}
\end{table}
\begin{table}[htb]
\centering
\begin{tabular}{c|cccc}
Model &MAD$\downarrow$  &  MSE$\downarrow$  &  Grad$\downarrow$ & dtSSD $\downarrow$  \\ \hline %\midrule[1.5pt]  %dtSSD\\ \hline %\midrule[1.5pt] &12.35
MODNet~\cite{sun2021modnet} &24.04 &15.53 &38.88 &12.35\\
RVM~\cite{lin2021robust}  &27.50 &21.31 &34.18 &17.16\\ %&17.16
%MSG-VIM MTRCNN  &60.84 &53.11 &44.86 \\ %17.65
MSG-VIM  &\textbf{20.30} &\textbf{13.91} &\textbf{22.03} &\textbf{9.77} \\
\end{tabular}
\vspace{-2mm}
\caption{Evaluation of video matting methods and MSG-VIM on VIM50 under video matting metrics. }
\vspace{-6mm}
%\textbf{Bold} numbers indicate best performance
%among all models.}
\label{tab:vim50_vm}
\end{table}

\subsection{Ablation Studies}
\vspace{-1mm}
\noindent \textbf{MSG-VIM Setup} 
We analyze the impact of the proposed model improvements of Section~\ref{sec:method}. The first line of Table~\ref{ablation_msgvim} is the baseline model, as described in Section~\ref{ssec:msg-vim}, on VIM50 under mask guidance obtained from MaskTrackRCNN, without any augmentation or temporal guidance. When we \revise{gradually} add our mixture of mask augmentations~(MMA), Temporal Mask Guidance~(TMG), and Temporal Feature Guidance~(TFG) to the baseline model, the performance gains are 9.15, 12.25 and 13.52 \wrt VIMQ over the baseline model, respectively. We refer to MSG-VIM as the model including all these improvements.
%\RHT{Highlight in percentage exemplary how much is the improvement.}
%To set up the baseline model for video instance matting, we first build a mask sequence guided video matting model, which takes video and mask sequences as inputs. Then, we apply training data with multiple human instances as foregrounds with separated target mask and reference mask sequence guidance as shown in the first line of Table~\ref{ablation_msgvim}. 
%We further add Mixed Mask Augmentation~(MMA), Temporal Mask Guidance~(TMG), and Temporal Feature Guidance~(TFG) to the baseline model, which gradually improve the performance on all metrics. Finally, the whole MSG-VIM model is built as shown in Figure~\ref{fig:architecture}.\\

\noindent \textbf{Mixture of Mask Augmentation} 
We further analyze the influence of the different methods involved in our proposed mixture of mask augmentations, as shown in Table~\ref{ablation_mma}. Without Mask Erase, Mask Merge and Mask Paste, the performance drops 6.86, 6.08 and 2.30 \wrt VIMQ, respectively. It shows that MSG-VIM benefits from each mask-oriented data augmentation to improve its robustness to inaccurate mask guidance from VIS methods. \\
\noindent \textbf{Temporal Guidance} 
Finally, we analyze in detail the impact of temporal mask guidance~(TMG) and temporal feature guidance~(TFG) as discussed in Section \ref{ssec:temporal_guidance}.
We use dtSSD as metric $\xi$, which evaluates the effectiveness of the temporal guidance.
The results presented in Table~\ref{ablation_tg} show that 
using TMG and TFG further improves both the tracking score and temporal consistency of alpha matte predictions to 18.41 VIMQ$_{\mathrm{dtssd}}$ and 19.32 VIMQ$_{\mathrm{dtssd}}$, justifying the relevance of these temporal guidance approaches. 

\noindent \textbf{Robustness Analysis}
We analyze the dependency of MSG-VIM on the accuracy of the mask sequence guidance input. We turn ground truth alpha mattes of VIM50 into masks via binarization and track the performance of MSG-VIM on the perturbed input as we add noise. As shown in Figure~\ref{fig:vis_mask_per}, when no noise is applied, both models achieve about the same accuracy. When the input data is perturbed, 
the performance drops for both models. However, the impact is less severe when temporal mask guidance, temporal feature guidance, and the mixture of mask augmentations are all used. For instance, when we apply Salt-and-Pepper noise to 25\% of the pixels of the mask guidance, the baseline model achieves a VIMQ$_{\mathrm{mse}}$ value of around 40\%, while for the full MSG-VIM model, the performance drops only slightly from 99\% to 91\%. It shows the effectiveness of our model improvements, which mitigate input errors due to the exploitation of temporal information. %These components thus lead to superior results.
\subsection{Comparisons to SOTA methods}
\vspace{-1mm}
To make comprehensive comparisons to other methods, \revise{we select state-of-the-art methods from three relevant categories: video instance segmentation, video matting, and mask-guided image matting.} We then evaluate these methods on the VIM50 benchmark as shown in Table~\ref{sota_iccv}. For video instance segmentation, we use the ResNet-50~\cite{he2016deep} based checkpoints of the two well-established VIS models: CNN-based MaskTrackRCNN~\cite{yang2019video} and the most recent state-of-the-art transformer-based SeqFormer~\cite{wu2021seqformer}. For video matting methods, \revise{considering that VIM50 does not provide trimap annotations and trimap-free inference is closer to real-world applications}, we select two state-of-the-art \revise{trimap-free} video matting models: MobileNetV3~\cite{howard2019searching} based MODNet~\cite{ke2020green} and RVM~\cite{lin2021robust} for evaluation. Since they do not decompose alpha mattes into corresponding instances, we perform the assignment into instances in the post-processing step. To this end, we binarize alpha mattes into masks, identify in each frame the connected components, and then link each component across time via overlap maximization using the Hungarian algorithm to form for each instance the alpha matte sequence. We also compare our method with mask-guided image matting methods MGMatting~\cite{yu2021mask} and InstMatt~\cite{sun2022human}. Since both methods are designed for image matting, we apply mask guidance frame-by-frame during inference. We re-train ResNet34-UNet~\cite{li2020natural} based MGMatting and InstMatt on our training set and evaluate the performance of both methods using MaskTrackRCNN and SeqFormer for the mask guidance. MGMatting and InstMatt successfully refine the mask predictions into instance-level alpha matte in each frame. Yet, MSG-VIM performs much better, especially on the metrics evaluating temporal consistency, \eg VIMQ$_{\mathrm{dtssd}}$. MSG-VIM also benefits from the more accurate mask guidance from SeqFormer \textit{without} re-training the MSG-VIM model. We also evaluate the inference speed of each model under a single A6000 with input size at $512 \times 288$ without extra optimization during inference.

\begin{figure}[tb]
\centering
\includegraphics[width=0.5\textwidth]{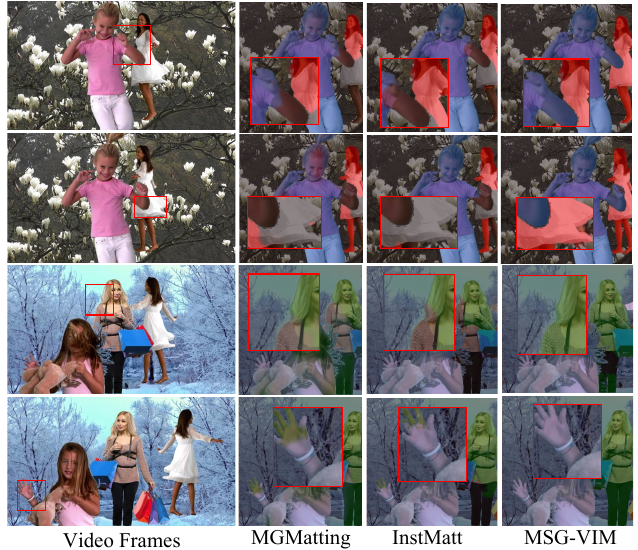}
\vspace{-6mm}
\caption{Qualitative comparisons of different methods with colored matting predictions of each instance in VIM50.}
%. We select challenging video frames from the VIM50 benchmark and visualize colored matting predictions of different models. Failure cases of other models are highlighted with red boxes.}
%\caption{\textbf{Qualitative comparisons between different models.} We select challenging video frames from the VIM50 benchmark and visualize predictions of different models. Failure cases of other models are highlighted with red boxes. Please zoom in for details.}
\vspace{-4mm}
\label{fig:visualize}
\end{figure}

\begin{figure}[tb]
\centering
\includegraphics[width=0.5\textwidth]{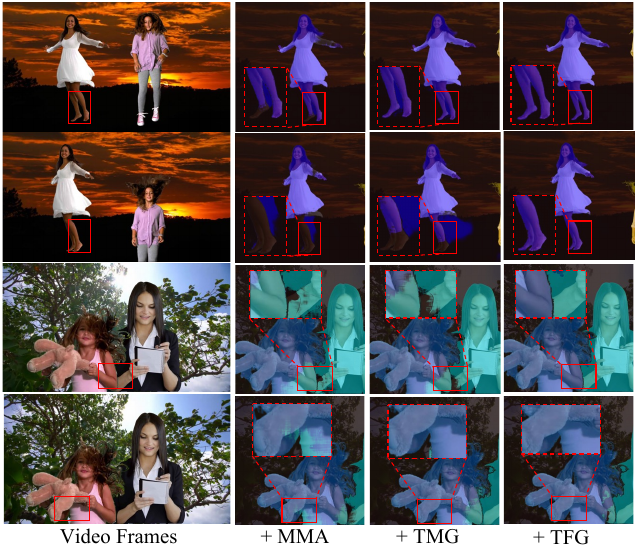}
\vspace{-6mm}
\caption{Qualitative comparisons of different modules with colored matting predictions of each instance in VIM50.}
\vspace{-7mm}
\label{fig:ablation}
\end{figure}

\begin{figure}[tb]
\centering
\includegraphics[width=0.49\textwidth]{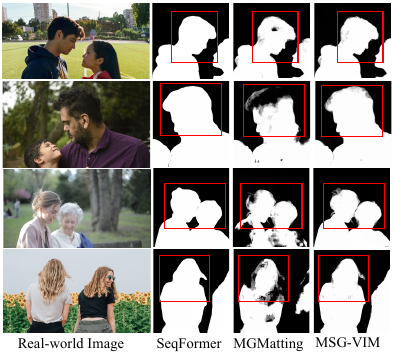}
\vspace{-6mm}
\caption{Qualitative comparisons on alpha matte quality of real-world data.}
\vspace{-6mm}
\label{fig:realworld}
\end{figure}

\subsection{Video Matting Extension}
\vspace{-1mm}
\label{ssec:videomatting}
\noindent \textbf{Video matting benchmark} While MSG-VIM is designed for video instance matting, it can also be applied to conventional video matting by merging alpha mattes of all instances into one alpha matte at each frame. We evaluate MSG-VIM on the high-resolution test set used in RVM~\cite{lin2021robust}.
The performance in Table~\ref{rvm_hr_testet} shows that MSG-VIM reaches better results compared to previous video matting works. It can thus be considered as high-quality VIM method, while still being useful for video matting, where it delivers state-of-the-art  results.

\noindent \textbf{Difficulty Analysis of VIM50} Any video instance matting dataset can be transformed into a video matting dataset by ignoring all instance-related information. Thus, to estimate the difficulty of VIM50, we evaluate state-of-the-art video matting methods  RVM~\cite{lin2021robust}, MODNet~\cite{sun2021modnet} as well as MSG-VIM on VIM50 from the video matting perspective. Comparing the same error metrics of Table~\ref{tab:vim50_vm} with Table~\ref{rvm_hr_testet} indicates that the VIM50 benchmark is indeed challenging, already on the video matting task. For the state-of-the-art video matting benchmark presented in RVM\cite{lin2021robust}, the MAD metric ranges between 6.47 (MSG-VIM) and 20.35 (BGMv2~\cite{lin2021real})  and the MSE error between 1.73 (MSG-VIM) and 14.26 (BGMv2). %In contrast to that, we observed MAD values between  20.3 (MSG-VIM) and 27.5 (MODNet) and MSE errors between 13.91 (MSG-VIM) and 21.31 (RVM) on VIM50.
\subsection{Qualitative Results}
\vspace{-1mm}
\noindent \textbf{VIM50 benchmark} We select frames of the VIM50 benchmark and visualize them in Figure~\ref{fig:visualize} and Figure~\ref{fig:ablation} with colored matting results of different methods and modules. While the mask guidance is inaccurate from MTRCNN, our method is able to correct most errors and produces more accurate alpha mattes than the results from competing methods. The errors are also \revise{gradually} reduced with the proposed modules \revise{cumulatively} added. 

\noindent \textbf{Real-world data} We further select the real-world images with multiple human instances from HIM2K~\cite{sun2022human}, and visualize the prediction of alpha mattes under SeqFormer, MGMatting, and MSG-VIM in Figure~\ref{fig:realworld}. 

\section{Conclusion}
\vspace{-1mm}
In this paper, we propose Video Instance Matting~(VIM), a new task aiming at estimating the alpha mattes of each instance at each frame of a video sequence. Furthermore, we establish the VIM50 benchmark and the VIMQ metric to evaluate the performance of different models on the new task. We propose MSG-VIM, a Mask Sequence Guided Video Instance Matting network, as a baseline model for the VIM50 benchmark. It benefits from our mixture of mask augmentations, temporal mask guidance, and temporal feature guidance. It outperforms previous methods by a large margin on our VIM50 benchmark and current methods on the video matting task.
We anticipate numerous new applications arising from the proposed task and that the presented dataset will drive research in this direction. In addition, research conducted on VIM may lead to advances in classical video matting, as our studies suggest, due to the great performance of MSG-VIM in this field.

%%%%%%%%% REFERENCES
{\small
\bibliographystyle{ieee_fullname}
\bibliography{egbib}
}

\newpage
%\title{Video Instance Matting}
\appendix
\section*{\Large{Appendix}}

\begin{figure}[tb]
\centering
 \includegraphics[width=0.5\textwidth]{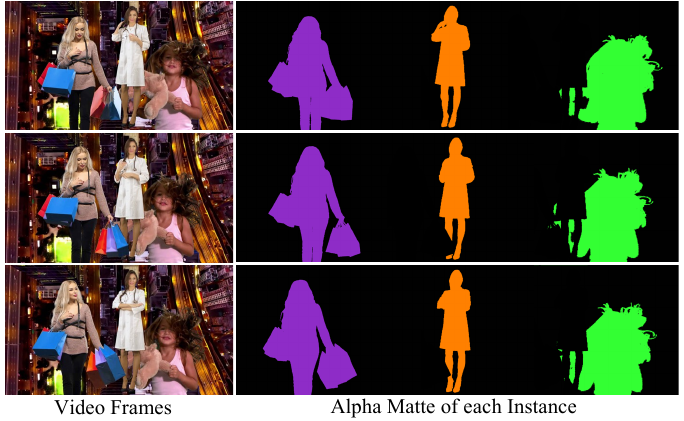}
\caption{Exemplary frames (first column) of the VIM50 benchmark with corresponding ground truth alpha mattes and instance information visualized in terms of the color encoding (second column).}
\vspace{-3mm}
\label{fig:sample}
\end{figure}

\begin{figure*}[tb]
\centering
 \includegraphics[width=1.0\textwidth]{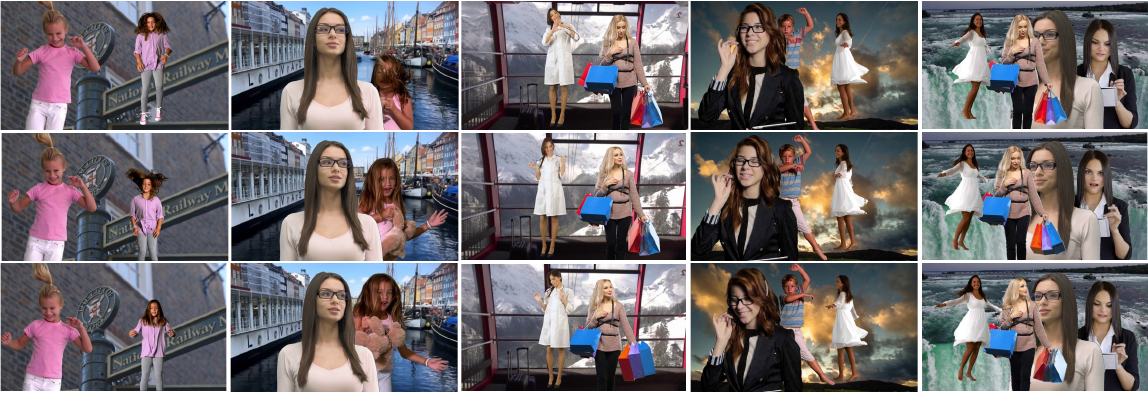}
\caption{Each column shows exemplary frames of one sequence of the VIM50 benchmark. Some frames contain heavy occlusions between persons, making it challenging for current VIS/VM/VIM methods to obtain accurate alpha matte predictions.}
\vspace{-3mm}
\label{fig:frames}
\end{figure*}

%%% add introduction here
This supplementary material elaborates on further aspects of our work regarding the benchmark VIM50 and the model MSG-VIM. 
In Appendix~\ref{appendix:benchmark}, we show additional video frames and ground truth data of the VIM50 test set. 
%
%Appendix~\ref{appendix:vim_for_matting} puts VIM50 into perspective with existing video matting datasets. 
Appendix~\ref{appendix:qualitative_results} shows more qualitative results of MSG-VIM. 
Additional details on the matting model architecture of MSG-VIM are provided in Appendix~\ref{appendix:model_architecture}.
Appendix~\ref{appendix:parameter_study} provides the parameter study we used to set the hyperparameters of MSG-VIM for the experiments presented in Section~\ref{sec:experiments}.

\section{VIM50 Benchmark}
\label{appendix:benchmark}
%%%%
To supplement the VIM50 samples presented in Section~\ref{ssec:vim50}, we show clips from five sequences of the VIM50 benchmark in Figure~\ref{fig:frames}. They depict two to four human instances as foreground objects with some frames containing heavy occlusions. 
Additional samples of the benchmark with corresponding individual ground truth alpha mattes are shown in Figure~\ref{fig:sample}. Ground truth alpha mattes belonging to the same person are colored consistently across video frames.

\begin{figure*}[tb]
\centering
 \includegraphics[width=0.99\textwidth]{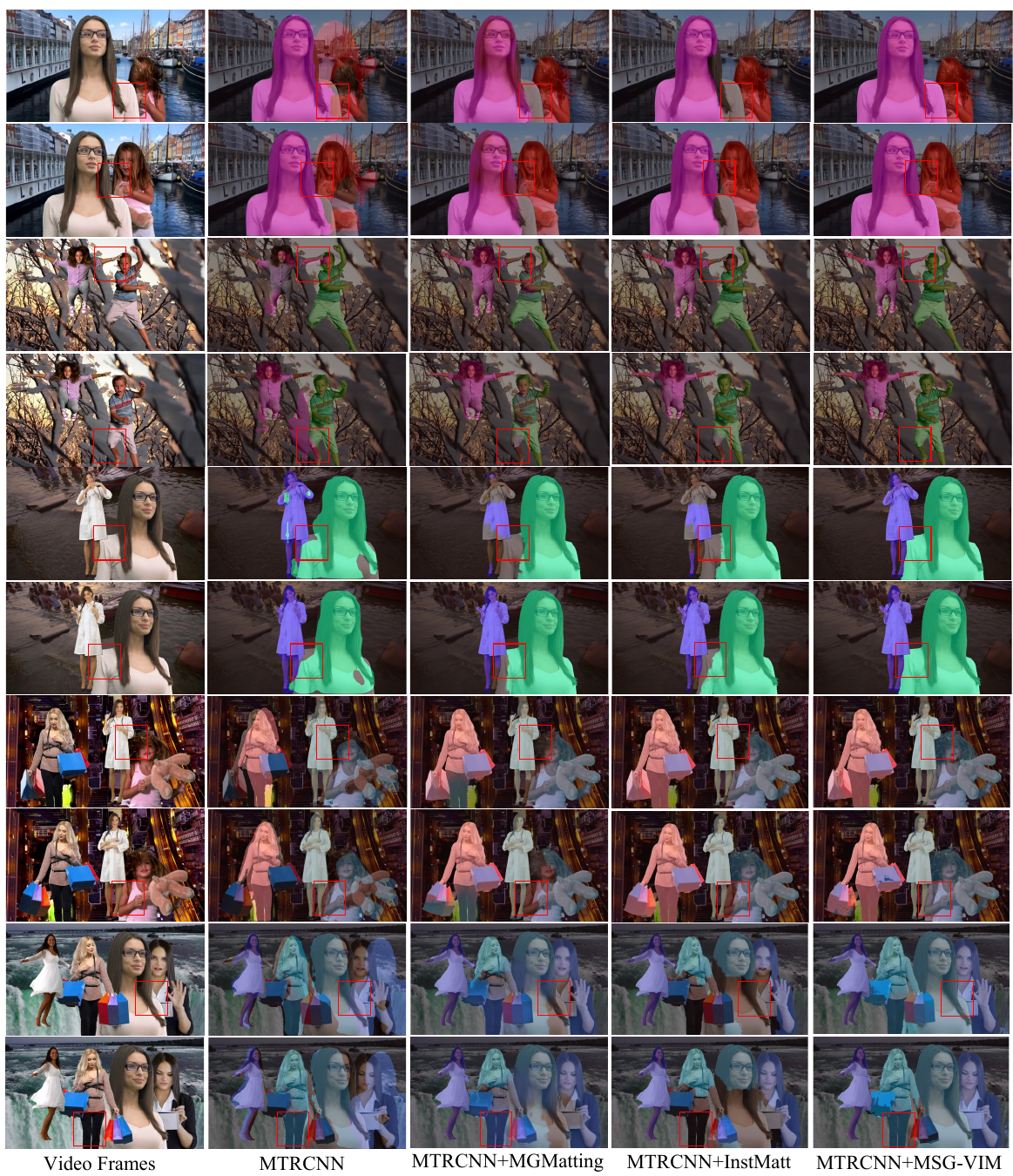}
 %\vspace*{-3mm}
\caption{Video instance matting results of different models on VIM50. For each row, the first column shows the input frame, column 2-5 show the matting result of the respective frame and method.
Difficult cases are highlighted with red boxes. Please zoom in for details. %\RHT{It sounds like cherry picking. We select frames where other methods perform worse and show only their errors.}
}
\vspace{-3mm}
\label{fig:vis_supp}
\end{figure*}

%\begin{figure*}[tb]
%\centering
% \includegraphics[width=1.0\textwidth]{figures/supp/fig_ablation_supp_v2.pdf}
%\vspace*{-3mm}
%\caption{Video instance matting results on VIM50, obtained using different components of VIM-MSG. For each row, the first column shows the input frame, column 2-5 show the matting result of the respective frame when the baseline model (Section~\ref{ssec:msg-vim}) and different additional components, described in Section~\ref{ssec:mix_aug} and Section~\ref{ssec:temporal_guidance}, are used.
%
%Qualitative comparisons between different module gradually added to the base model on selected video frames in VIM50. 
%Failure cases of the ablated models are highlighted with red boxes. Please zoom in for details.}
%\vspace{-1mm}
%\label{fig:vis_ablation_supp}
%\end{figure*}

\section{Additional Qualitative Results}
\label{appendix:qualitative_results}
We visualize in Figure~\ref{fig:vis_supp} additional qualitative results on selected video frames of VIM50. Similar to the qualitative comparison conducted for Figure~\ref{fig:visualize}, we use mask guidance from MaskTrackRCNN~\cite{yang2019video} and compare MSG-VIM with 
our transformation of MGMatting and InstMatt to video instance matting.
%the reproduced MGMatting and InstMatt methods. %We highlight the failure cases of these models in red boxes on the same frame to show the better performance of MSG-VIM qualitatively.
The results show visually a significant advantage of MSG-VIM over the baseline methods. 

%We further demonstrate qualitatively the impact of each proposed module (see Section~\ref{ssec:mix_aug} and Section~\ref{ssec:temporal_guidance}) over the baseline setup of MSG-VIM (see Section~\ref{ssec:msg-vim}).
%
%produce qualitative results of each module that we proposed during the setup of MSG-VIM in Section~\ref{ssec:msg-vim}. 
%As shown in Figure~\ref{fig:vis_ablation_supp}, our baseline model is not sufficient for high-quality VIM results. We observe severe errors, especially at the limbs. By using a mixture of mask augmentations~(MMA), temporal mask guidance~(TMG), and temporal feature guidance~(TFG), the results show that the error cases are gradually prevented, leading to a significantly more accurate alpha matte prediction for each instance. 

%As shown in Figure~\ref{fig:vis_ablation_supp}, our base model is built without the mixture of mask augmentation~(MMA), temporal modeling guidance~(TMG), and temporal feature guidance~(TFG). When we gradually add MMA, TMG, and TFG to the Base Model, the qualitative results show that the failure cases are cumulatively refined to more accurate alpha matte predictions of each instance.

\begin{table}[htb]
\centering
\resizebox{0.5\textwidth}{!}{
\begin{tabular}{c|c|c} \hline
Layers     &Output Size &MSG-VIM \\  \hline
Convolution &256 $\times$ 256
& $\bigr[$ \begin{tabular}{c}
        3$\times$3 Conv \\
        \end{tabular} $\bigr]$ $\times$ 2 \\ \hline
ResNet Block (1) &128 $\times$ 128
& $\Bigr[$ \begin{tabular}{c}
        3$\times$3 Conv \\
        3$\times$3 Conv \\
        \end{tabular} $\Bigr]$ $\times$ 3 \\ \hline
ResNet Block (2) &64 $\times$ 64
& $\Bigr[$ \begin{tabular}{c}
        3$\times$3 Conv \\
        3$\times$3 Conv \\
        \end{tabular} $\Bigr]$ $\times$ 4 \\ \hline
ResNet Block (3) &32 $\times$ 32
& $\Bigr[$ \begin{tabular}{c}
        3$\times$3 Conv \\
        3$\times$3 Conv \\
        \end{tabular} $\Bigr]$ $\times$ 4 \\ \hline
ResNet Block (4) &16 $\times$ 16
& $\Bigr[$ \begin{tabular}{c}
        3$\times$3 Conv \\
        3$\times$3 Conv \\
        \end{tabular} $\Bigr]$ $\times$ 2 \\ \hline
ASPP &16 $\times$ 16
& dilations = [1, 2, 4, 8] \\ \hline
Upsample Block (1) &32 $\times$ 32
& $\Bigr[$ \begin{tabular}{c}
        3$\times$3 Conv \\
        3$\times$3 Conv \\
        \end{tabular} $\Bigr]$ $\times$ 2 \\ \hline
Upsample Block (2) &64 $\times$ 64
& $\Bigr[$ \begin{tabular}{c}
        3$\times$3 Conv \\
        3$\times$3 Conv \\
        \end{tabular} $\Bigr]$ $\times$ 3 \\ \hline
Upsample Block (3) &128 $\times$ 128
& $\Bigr[$ \begin{tabular}{c}
        3$\times$3 Conv \\
        3$\times$3 Conv \\
        \end{tabular} $\Bigr]$ $\times$ 3 \\ \hline
Upsample Block (4) &256 $\times$ 256
& $\Bigr[$ \begin{tabular}{c}
        3$\times$3 Conv \\
        3$\times$3 Conv \\
        \end{tabular} $\Bigr]$ $\times$ 2 \\ \hline
ConvRNN &256 $\times$ 256
&TFG \\ \hline
Deconvolution &512 $\times$ 512
& 4$\times$4 Deconv \\ \hline
\end{tabular}}
\caption{The detailed architecture of the matting network $U$ used in MSG-VIM. TFG denotes temporal feature guidance presented in Section~\ref{ssec:temporal_guidance}. Pooling layers and normalization layers are omitted for simplicity.} %\RHT{Looks like the TFG Layer is doing only a convoltion. We should have TFG on the right side. Or RNN?}}
\label{tab:model}
\end{table}

\section{Matting Architecture}
\label{appendix:model_architecture}
In Table~\ref{tab:model} we present additional details on the architecture of the encoder-decoder-based matting network MSG-VIM (compare Section~\ref{ssec:msg-vim}). The encoder is adopted from the modified ResNet-34~\cite{yu2021mask}. Each of its ResNet blocks contains consecutive $3\times3$ convolution layers with a final average pooling layer to downsample the feature maps. The decoder has multiple upsample blocks. Each one consists of consecutive $3\times3$ convolution layers with a final 2D nearest neighbor upsampling layer. The temporal feature guidance~(TFP) module, which is implemented using a Convolution-based RNN~(ConvRNN) network, is applied to the second largest feature map with a resolution of $256 \times 256$ in the decoder, as described in Section~\ref{ssec:temporal_guidance}. At the first frame $t=0$, we initialize the internal state via $h_{0} = \mathrm{tanh} ( \mathrm{Conv}(F_0^{S}) )$. After this stage, a $4\times4$ deconvolution layer with stride 2 is used to upsample the feature map to size $512 \times 512$ for the final predictions of the alpha mattes.

\begin{table*}[tb]
\centering
\resizebox{1.0\textwidth}{!}{
\begin{tabular}{c|llllllll}
 Length of Chunk  &  RQ$\uparrow$ & TQ$\uparrow$ & MQ$_{\mathrm{mse}}$ $\uparrow$ & VIMQ$_{\mathrm{mse}}$ $\uparrow$ & MQ$_{\mathrm{mad}}$ $\uparrow$ & VIMQ$_{\mathrm{mad}}$ $\uparrow$ & MQ$_{\mathrm{dtssd}}$ $\uparrow$ & VIMQ$_{\mathrm{dtssd}}$ $\uparrow$\\ \hline
t = 1  &72.12 &92.03 &54.74 &36.33 &39.10 &25.95 &27.02 &17.93  \\ \hline
t = 5  &72.26 &\textbf{93.27} &56.15 &38.06 &40.31 &27.32 &28.36 &19.22 \\ \hline
t = 10 &\textbf{72.72} &93.17 &\textbf{56.52} &\textbf{38.29}  &\textbf{40.49} &\textbf{27.43} &\textbf{28.51} &\textbf{19.32}\\ \hline
\end{tabular}}
\caption{Analysis on the chunk length used during inference of MSG-VIM with mask sequence guidance from MaskTrackRCNN~\cite{yang2019video}. \textbf{Bold} numbers indicate best performance
among all models.}
\vspace{-2mm}
\label{ablation_inference_frames}
\end{table*}

\section{Parameter Study: Chunk Length}
\label{appendix:parameter_study}
%\subsection{Chunk Length}
Given a video sequence of length $T$, we split the input (video and mask guidance) into chunks of $t$ consecutive frames for inference. 
%
%
%During inference of the MSG-VIM model, given a video sequence of $T$ frames with mask guidance, we first split the video and mask sequence guidance into chunks, each consisting of $t$ frames. 
Then, we run inference on each chunk independently and concatenate results together. Identity  information across chunks are maintained from the underlying mask sequence generator, \eg MaskTrackRCNN. 
%

%For each chunk, we send the video and mask guidance frames with length $t$ to the MSG-VIM model, which returns the alpha matte of the target instance with length $t$. 
In Table~\ref{ablation_inference_frames}, we analyze the impact of the video length that is processed during the inference of one chunk. 
The results show that the performance of the model improves with longer length $t$. The proposed temporal feature guidance module is thus an effective approach to exploit temporal information.
Accordingly, we have set $t=10$ in all experiments presented in Section~\ref{sec:experiments}. Note that we could not process chunks with a length larger than $t=10$ due to memory limitations.

\end{document}